\documentclass[letterpaper, 10 pt, conference]{ieeeconf}  

\IEEEoverridecommandlockouts                                   

\usepackage{amsmath}
\usepackage{nicefrac}
\usepackage{units}
\usepackage[pdftex]{graphicx}
\usepackage{import}
\usepackage{xcolor}
\usepackage{colortbl}
\usepackage{caption}
\usepackage{cite}
\title{\LARGE \bf
    Reducing Warning Errors in Driver Support with \\ Personalized Risk Maps
}

\author{Tim Puphal$^{1,2}$, Ryohei Hirano$^3$, Takayuki Kawabuchi$^3$, Akihito Kimata$^3$ and Julian Eggert$^1$
\thanks{$^1$ Honda Research Institute Europe, Carl-Legien-Str. 30, 63073 Offenbach, Germany. Email: {\tt\footnotesize \{firstname.lastname\}@honda-ri.de}}
\thanks{$^2$ Honda Research Institute Japan, 8-1 Honcho, Wako, 351-0114 Saitamam Japan. Email: {\tt\footnotesize \{tim.puphal\}@jp.honda-ri.com}}
\thanks{
$^3$ Honda R$\&$D Co., Ltd. 4630 Shimotakanezawa, 321-3393 Tochigi, Japan. Email: {\tt\footnotesize \{firstname$\_$lastname\}@jp.honda}}
}

\begin{document}

\maketitle
\thispagestyle{empty}
\pagestyle{empty}

\begin{abstract}
We consider the problem of human-focused driver support. State-of-the-art personalization concepts allow to estimate parameters for vehicle control systems or driver models. However, there are currently few approaches proposed that use personalized models and evaluate the effectiveness in the form of general risk warning. In this paper, we therefore propose a warning system that estimates a personalized risk factor for the given driver based on the driver's behavior. The system afterwards is able to adapt the warning signal with personalized Risk Maps. In experiments, we show examples for longitudinal following and intersection scenarios in which the novel warning system can effectively reduce false negative errors and false positive errors compared to a baseline approach which does not use personalized driver considerations. This underlines the potential of personalization for reducing warning errors in risk warning and driver support.
\end{abstract}

\section{Introduction}
Driver support systems are helping nowadays drivers in many everyday driving situations. In current vehicles, advanced systems like adaptive cruise control systems, collision mitigation systems or parking assistance systems have been successfully introduced \cite{bengler2014}. However, although drivers can manually change the system settings to improve their experience of these systems, there are still instances where support is applied or warnings are triggered even though the driver prefers not to have driver support. To solve this problem, research is increasingly focusing on personalization concepts which automatically adapt the system to each driver. As an example, a method introduced in \cite{wang2013} estimates parameters of a vehicle control system based on the driver behavior during car following. 

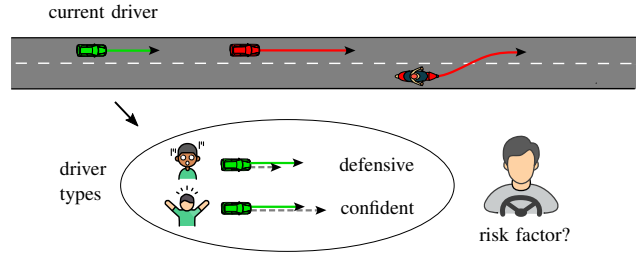
\begin{figure}[t!]
  \centering
  \vspace*{0.17cm}
  \resizebox{0.96\linewidth}{!}{
\begingroup%
  \makeatletter%
  \providecommand\color[2][]{%
    \errmessage{(Inkscape) Color is used for the text in Inkscape, but the package 'color.sty' is not loaded}%
    \renewcommand\color[2][]{}%
  }%
  \providecommand\transparent[1]{%
    \errmessage{(Inkscape) Transparency is used (non-zero) for the text in Inkscape, but the package 'transparent.sty' is not loaded}%
    \renewcommand\transparent[1]{}%
  }%
  \providecommand\rotatebox[2]{#2}%
  \newcommand*\fsize{\dimexpr\f@size pt\relax}%
  \newcommand*\lineheight[1]{\fontsize{\fsize}{#1\fsize}\selectfont}%
  \ifx\svgwidth\undefined%
    \setlength{\unitlength}{315.21773499bp}%
    \ifx\svgscale\undefined%
      \relax%
    \else%
      \setlength{\unitlength}{\unitlength * \real{\svgscale}}%
    \fi%
  \else%
    \setlength{\unitlength}{\svgwidth}%
  \fi%
  \global\let\svgwidth\undefined%
  \global\let\svgscale\undefined%
  \makeatother%
  \begin{picture}(1,0.40648339)%
    \lineheight{1}%
    \setlength\tabcolsep{0pt}%
    \put(0.52595737,0.1330864){\color[rgb]{0,0,0}\makebox(0,0)[lt]{\lineheight{1.25}\smash{\begin{tabular}[t]{l}defensive\end{tabular}}}}%
    \put(0.52815676,0.0650746){\color[rgb]{0,0,0}\makebox(0,0)[lt]{\lineheight{1.25}\smash{\begin{tabular}[t]{l}confident\end{tabular}}}}%
    \put(0.05970465,0.37933239){\color[rgb]{0,0,0}\makebox(0,0)[lt]{\lineheight{1.25}\smash{\begin{tabular}[t]{l}current driver\end{tabular}}}}%
    \put(0.077826,0.12682778){\color[rgb]{0,0,0}\makebox(0,0)[lt]{\lineheight{1.25}\smash{\begin{tabular}[t]{l}driver\\types\end{tabular}}}}%
    \put(0.75037696,0.02105799){\color[rgb]{0,0,0}\makebox(0,0)[lt]{\lineheight{1.25}\smash{\begin{tabular}[t]{l}risk factor?\end{tabular}}}}%
    \put(0,0){\includegraphics[width=\unitlength,page=1]{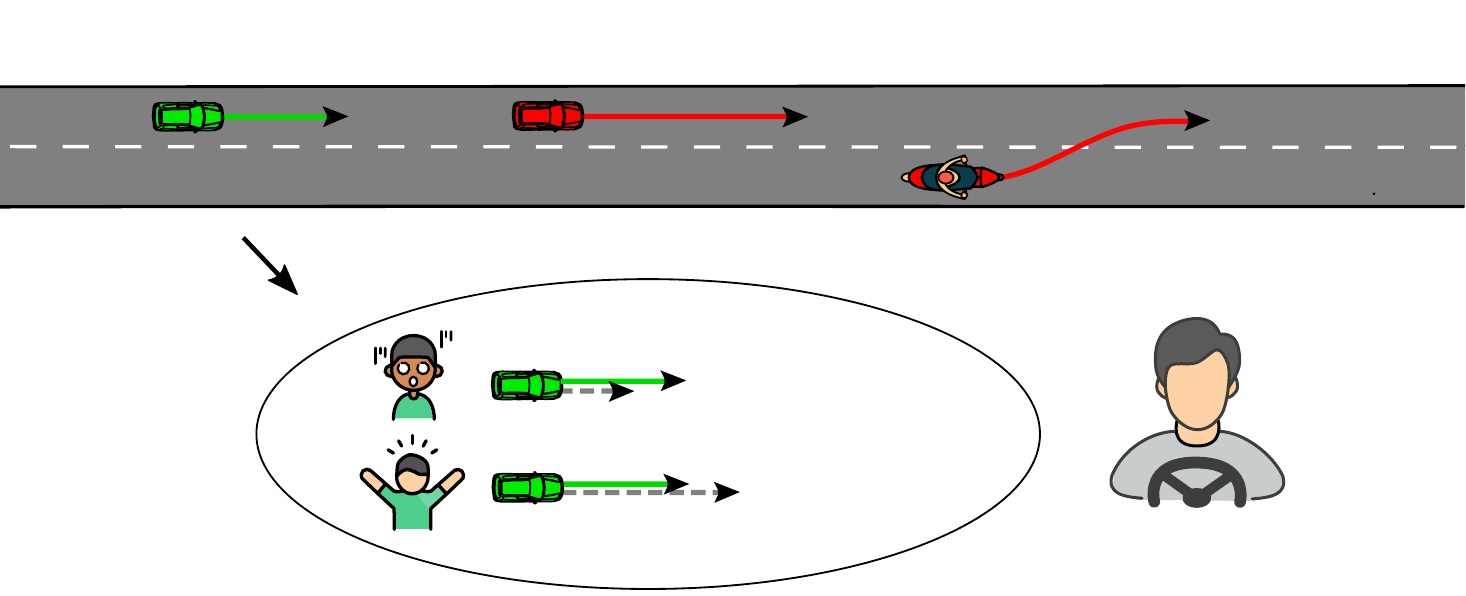}}%
  \end{picture}%
\endgroup%
}
  \vspace*{0.06cm}
  \caption[]{The image shows a driving situation example of a dynamic car following in which the driver in the green car can react differently depending on the driver type. In this paper, we propose a driver support in the form of warning that can reduce driver warning errors by using these driver types in personalized Risk Maps.}
  \vspace*{-0.14cm}
  \label{fig:intro}
\end{figure}

Fig. \ref{fig:intro} shows an example of a dynamic driving situation in which such personalization can improve driver support. In this situation, a red car brakes because of a motorcycle changing lanes. The driver of the green car needs to react now on the braking car with the reaction varying according to the driver type. For example, the driver could brake if the driver is defensive, or keep on driving or accelerating if the driver is confident and believes to have enough time to react later. Using these driver types, the driver support and driver warning can be improved. For instance, by detecting that the driver is defensive allows to decide if a driver prefers a warning or not. In effect, this enables to reduce warning errors in the system. 
\vspace{0.3cm}

In this paper, we propose a system that warns the driver by using driver types in the form of a personalized risk factor. The system consists of an estimation module that estimates the driver type based on a personalized risk factor and a warning concept that uses the factor in personalized Risk Maps. Here, personalized Risk Maps can visualize the risk that the individual driver predicts for different future driver behaviors. We show in experiments that the personalized modeling in Risk Maps effectively improves driver support. Examples of results for longitudinal following and intersection scenarios demonstrate that the warning system is able to reduce false negative errors and false positive errors compared to a baseline which does not use personalized models.

\subsection{Related Work}
In the past, we have already shown that the human state (e.g., the driver's gaze or awareness) sensed by cameras or other sensors can help to develop perceived Risk Maps and to warn of driver perception errors \cite{puphal2023}. While sensing the human state is one way to consider the human driver in driver support, this is not the focus of this paper. 

Instead, in this paper, we explore personalization of driver support by using motion data of the driver's vehicle. The system can thus be adapted to the general driving behavior type of the driver to improve driver support. In \cite{ingelder2016}, motion data was shown to correlate with the driving style, which underlines the potential of this alternative strategy. 
In this context, the authors of \cite{orth2017} estimated driver-specific critical intersection merging gaps from driving data with machine learning. The authors of \cite{kreutz2022}, estimated reaction times for the driver model called Intelligent Driver Model (IDM), and the authors of the paper \cite{kolekar2020} adapted parameters for a driver's risk field to different driving situations, such as curve driving, and could safely handle the scenarios when coupling the system to a vehicle controller. 

For a comparison of further personalization concepts, we refer the reader to Hasenj\"ager et al. \cite{hasenjaeger2017} where trends and limitations of state-of-the-art personalized systems are discussed.
While there are many approaches that estimate parameters of vehicle control systems or driver models, such as 
\cite{ingelder2016,orth2017,kreutz2022,kolekar2020}, there are only few approaches proposed that use personalization and evaluate the effectiveness in the form of general risk warning. One distinctive example is \cite{wang2018}, which is applied for lane departure warning, and \cite{panou2018}, which is applied for frontal collision warning. However, these approaches use simple time-based risk models and are limited in the scenarios they can handle.

\subsection{Contribution} 
In this paper, we build upon related work and propose to use personalization for general risk warning systems. The paper evaluates the extent of the effectiveness of personalization for 
driver warning. In detail, the contributions are:
\begin{enumerate}
    \item We propose an estimation module for a personalized risk factor. The module allows to classify drivers into  defensive, normal and confident driver types. 
    \item We present a warning system that uses the estimated risk factor to form a personalized Risk Maps. We can thus adapt the warning signal to reduce warning errors, such as false negative and false positive errors. 
\end{enumerate}    
In comparison to previous work, the improvement by personalization can be achieved both for car-following and intersection scenarios and is not limited to one specific scenario. The results highlight the potential of personalization for general risk warning systems and driver support.

The remainder of the paper is structured accordingly. In the next Section \ref{sec:human_factors}, the estimation module for the personalized risk factor is explained. Afterwards, we describe the novel warning system with personalized Risk Maps that uses the personalized risk factor in Section \ref{sec:warning_system}. 
Experiments giving examples for the performance of the risk factor estimation module and the novel warning system are shown in Section \ref{sec:experiments} and Section \ref{sec:conclusion} outlines the conclusion and outlook for possible future work of this paper.

\vspace{0.1cm}
\section{Risk Factor Estimation}
\label{sec:human_factors}

The focus of this paper lies in the personalization of driver support in the form of driver risk warning. In this section, we will therefore explain the proposed personalization based on a driver's risk factor estimation. 

As mentioned in the introduction, one way to characterize distinct driver types is to utilize a personalized risk factor. There are, on the one hand, defensive driver types that prefer to keep distance to other vehicles. Since defensive drivers cannot estimate the behavior of other vehicles well, the driver sees uncertainty in the behavior and, in turn, predicts high risk and prefers to keep distance to other vehicles. On the other hand, there are confident driver types that drive close to other vehicles. In comparison to defensive drivers, confident driver types are good at estimating the behavior of vehicles. These drivers perceive reduced behavior uncertainty and risks and hence prefer to keep short distances while driving. 

In this manner, we define the personalized risk factor by how well the driver predicts vehicle behaviors into the future. The risk factor includes the amount of assumed uncertainty in the behavior of other vehicles and the uncertainty in the behavior of the own vehicle. For the risk factor estimation, we make use here of the behavior planner Risk Maps which we have previously proposed \cite{puphal2022}. Risk Maps is a planner that models risks for different possible driver behaviors over the future time. The planner contains the vehicle behavior uncertainties as parameters that change the planned behavior. In contrast to work \cite{puphal2022}, in which the uncertainty parameters are fixed, we personalize the parameters to driver types.

\subsection{Estimation Module}
\label{sec:alpha_estimation}
\begin{figure}[t!]
  \centering
  \vspace{0.22cm}
  \resizebox{0.96\linewidth}{!}{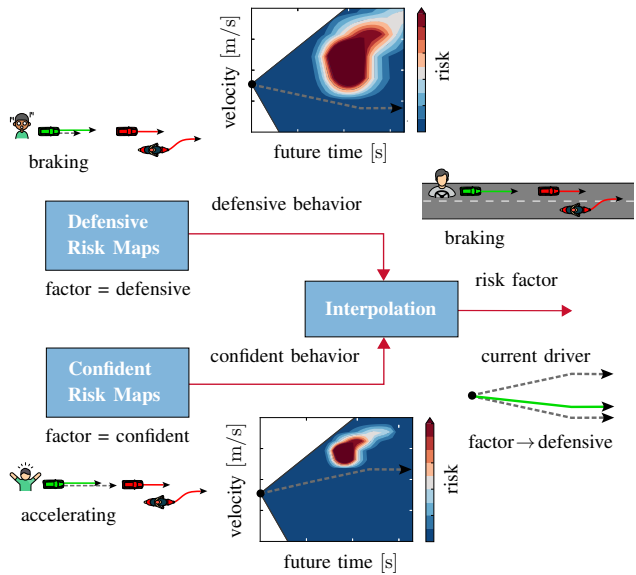}
  \vspace*{0.12cm}
  \caption[]{For the risk factor estimation, we plan driver behaviors by using the behavior planner Risk Maps with a defensive risk factor parametrization and a confident risk factor parametrization. By comparing the current driver behavior with the behavior plans, we can estimate the risk factor.}
  \vspace*{-0.03cm}
  \label{fig:risk_factor_estimation}
\end{figure}

A block diagram explaining the estimation module for the risk factor is shown in Fig. \ref{fig:risk_factor_estimation}. In the estimation module, we apply the following computational steps. First, we plan driver behaviors by using the behavior planner Risk Maps both for a defensive driver with the risk factor parametrization $\alpha_{\text{def}}$ and for a confident driver with the risk factor parametrization $\alpha_{\text{conf}}$. 
In this context, the defensive and confident parametrization represent the two extremes of the risk factor, which means that we set one small value and one large value for the risk factor. In this way, the current driver behavior will lie in between the found behavior plans of the defensive and confident Risk Maps. 

In a second step, we then estimate the personalized risk factor of the driver $\alpha$ by interpolating with the current driver behavior within the found behavior plans. The interpolation is given by 
\vspace*{-0.05cm}
\begin{equation}
\alpha = \alpha_{\text{conf}} + (\alpha_{\text{def}}-\alpha_{\text{conf}}) *\frac{a-a_{\text{conf}}}{a_{\text{def}}-a_{\text{conf}}}, 
\label{eq:interpolation}
\end{equation}
in which the behavior plans from the Risk Maps planner are parametrized by acceleration values $a_{\text{def}}$ and $a_{\text{conf}}$, and the driver behavior is given by the applied driver acceleration value $a$. In the actual implementation, we further include an interpolation using sigmoid functions because of the non-linear relationship of the risk factor and driver behavior is not linear in reality. One can also choose other interpolations, such as quadratic interpolation or others.

In order to plan an acceleration behavior, the planner Risk Maps generates several acceleration profiles $a^h$ for the driver that are composed of $h$ different acceleration behaviors and deceleration behaviors. The optimal driving behavior for the driver is the acceleration profile $a_{\text{opt}}$ among these generated behaviors that has the lowest driving costs $C$. The costs $C$ consist therein of future driving risks $R$, driving utility $U$ and comfort costs~$O$. We accordingly write 
\begin{align}
a_{\text{opt}} = \text{argmin}_{h} \{C(a^h)\}, \\
\text{with } C(a^h) = R(a^h) - U(a^h) &+ O(a^h), 
\end{align} 
for the planning formulation of Risk Maps. As an example, an optimal behavior for the driver could be to brake to follow another vehicle with a desired distance and afterwards to keep constant velocity for the rest of the planning time. The acceleration profile includes in this formulation also a time duration to reach a target velocity. 

A plot of defensive and confident Risk Maps that visually explains the behavior planning approach is shown in Fig. \ref{fig:risk_factor_estimation}. The plot allows to visualize predicted risks for all generated driver behaviors in the velocity-time space, whereby the planner chooses a behavior that avoids, amongst others, the risk spots colored in red. For the example of the dynamic car-following scenario, the defensive planner therefore finds a braking behavior with decreasing velocity that avoids the risk spot (see descending gray curve in the plot of defensive Risk Maps). In contrast, the confident planner finds an accelerating behavior with increasing velocity (see ascending gray curve in the plot of confident Risk Maps). Since the actual driver is braking, as shown with the green driver behavior curve in the figure, defensive Risk Maps is closer to the driver behavior. Eventually, the estimation module estimates the risk factor in this way to be close to a defensive risk factor. 

\vspace{0.05cm}
\section{Warning with Personalized Risk Maps}
\label{sec:warning_system}
In the last section, we described personalization using a personalized risk factor and we showed how this risk factor can be estimated by using the behavior planner Risk Maps. We target to apply the risk factor to improve driver warning and driver support. In this section, we therefore introduce a novel warning system that uses the risk factor in a risk model to form personalized Risk Maps. Furthermore, we present an adaptation for the warning signal output based on personalization. 

\subsection{Risk Model in Risk Maps}

The proposed warning system of this paper is shown in Fig. \ref{fig:warning_system}. The depicted personalized Risk Maps has as an input the vehicle states and the personalized risk factor. The used risk model in personalized Risk Maps represents here the Gaussian method from the work of \cite{puphal2022}. This method models two-dimensional Gaussian distributions around the vehicles over the future time. For the example of a driver encountering another vehicle, we model an uncertainty for the driver with $\mathbf{\Sigma}_{1}(t+s)$ and an uncertainty for the given other vehicle with $\mathbf{\Sigma}_{2}(t+s)$ that start at the current time~$t$ and are growing over the future time $s$. In the uncertainties, the risk factor is constraining the maximum assumed value for the driver and other vehicle.

In order to compute the collision risk for the driver for one chosen driver behavior at future times $s$, the Gaussian method takes the spatial overlap between the Gaussians of the driver vehicle and the other vehicle which are predicted at mean positions $\boldsymbol{\mu}_{1}(t+s)$ and $\boldsymbol{\mu}_{2}(t+s)$. As a result, we can define the driver risk as
\begin{align}
r(t+s; \alpha) \sim \operatorname{det}|2\pi&(\mathbf{\Sigma}_{1}(\alpha)+\mathbf{\Sigma}_{2}(\alpha))|^{-\frac{1}{2}} * \nonumber \\ \exp\{-\frac{1}{2}
(\boldsymbol{\mu}_{2}-\boldsymbol{\mu}_{1})^T(\mathbf{\Sigma}_{1}(\alpha)&+\mathbf{\Sigma}_{2}(\alpha))^{-1}
(\boldsymbol{\mu}_{2}-\boldsymbol{\mu}_{1})\}.   
\label{eq:prodgauss}
\end{align}
The equation denotes the uncertainty dependencies to the risk factor~$\alpha$ with $\mathbf{\Sigma}_1(\alpha)$ and $\mathbf{\Sigma}_2(\alpha)$. For better readability, we did not write out the dependency of the mean positions and uncertainties on the prediction time $t+s$. Additionally, we did not consider the collision severity in the risk for warning but we only used the collision probability. The severity can, however, be seamlessly integrated.

In the Gaussian method, the growth of the uncertainties in the form of Gaussian distributions included in the parameters $\mathbf{\Sigma}_{1,2}$ can be modeled with different strategies over the future time. For the risk warning system of this paper, we made the growth conditional on the vehicle velocity and took the maximum assumed uncertainty from the risk factor only for the longitudinal movement component. Here, the maximum values were set differently based on the risk factor for the ego driver and the other vehicles.

Fig. \ref{fig:warning_system} shows an example plot of personalized Risk Maps which visualizes the risk for different assumed driver behaviors. The plot shows the personalized risk and thus shows how the driver perceives the driving situation (``driver'' view). In the example of the figure, the driver sees high risk when braking (small velocity in lower plot area), and when driving with constant velocity (middle plot area). In contrast, the driver sees low risk when accelerating (high velocity in upper plot area). The personalized risk factor changes here the size of the red risk spot. A larger risk factor leads in general to larger assumed maximum uncertainties and therefore into a larger risk spot area.

\subsection{Adapting Warning Signal}

Based on personalized Risk Maps, we can retrieve the warning signal for the driver. Since we should warn the driver if the current behavior has high risk, we use the constant velocity driver behavior in the personalized Risk Maps plot to infer a warning signal. Concretely, the risk over all future times $r(t+s;\alpha)$ is integrated into a single scalar risk value for the current time. Here, we additionally model a Poisson distribution to rate risks less that occur later in future time. For details of the risk model and the modeling of the Poisson distribution with a so-called survival analysis, we refer the reader to our work \cite{puphal2022}. 

\begin{figure}[t!]
  \centering
  \vspace*{0.06cm}
  \resizebox{0.96\linewidth}{!}{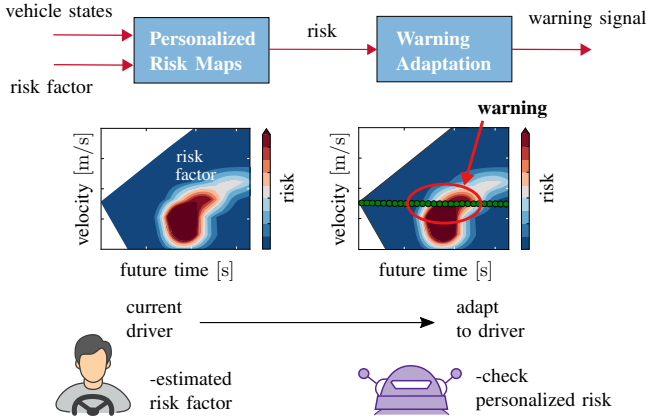}
  \vspace{-0.03cm}
  \caption[]{The warning system consists of Risk Maps that uses the risk factor to create a personalized Risk Maps and a warning adaptation module that uses a personalized weight to the risk value to obtain the warning signal.} 
  \label{fig:warning_system}
\end{figure}

Fig. \ref{fig:warning_system} highlights the constant driver behavior prediction in the personalized Risk Maps plot. The behavior enters the red risk spot and the warning system issues a warning. The warning system takes here the personalized risk factor $\alpha$ into account and the final risk value represents $R(\alpha)$.

In the warning adaptation module, we moreover incorporate a personalized weight onto the risk value to further adapt the warning system to the driver. We write 
\begin{equation}
W(t) = w_{\alpha} * R(\alpha)
\label{eq:warning_signal}
\end{equation}
for the warning signal at current time $t$. The risk weight~$w_{\alpha}$ is dependent on the personalized risk factor $\alpha$ based on a linear function that goes from $w_{\text{def}} > 1$ for a defensive driver over $w_{\text{norm}}=1$ for a normal driver to $w_{\text{conf}} < 1$ for a confident driver. The defensive and confident driver accordingly either overestimates or underestimates the risk based on the risk weight.

The risk factor $\alpha$ and risk weight $w_{\alpha}$ have different purposes for personalizing the driver warning.
While the risk factor allows to adapt the size of the risk spot and affects for which behavior the driver predicts risk, the weight influences how strong the driver predicts risk for one behavior and changes the magnitude or severity level of the risk spot.
As a comparison, warning systems without personalization would only use the risk value resulting from constant velocity behavior or other predictions from the driver with one general parametrization.

\section{Experiments}
\label{sec:experiments}

The following section describes the experiments that show the improved driver support using the presented warning system of personalized Risk Maps. Here, simulations were done with an own driving simulator written in python. The simulator allows to set a fixed driver type for the driver of one vehicle. We used the behavior planner Risk Maps with this driver type parametrization to simulate the driver and applied the proposed warning system of this paper in parallel to this simulation. 

The section is divided into three parts. In the first part, we show the performance of the personalized risk factor estimation for the different driver types of defensive, normal and confident drivers. 
In the second part, we use the averaged risk factor values and apply personalized Risk Maps to support the driver. Examples in which personalization allows to reduce warning errors and examples in which personalization does not exhibit improvements are shown. In the last part, we discuss the limitations of the conducted simulations.

\begin{figure*}[t!]
  \centering
  \vspace*{0.17cm}
  \resizebox{0.97\linewidth}{!}{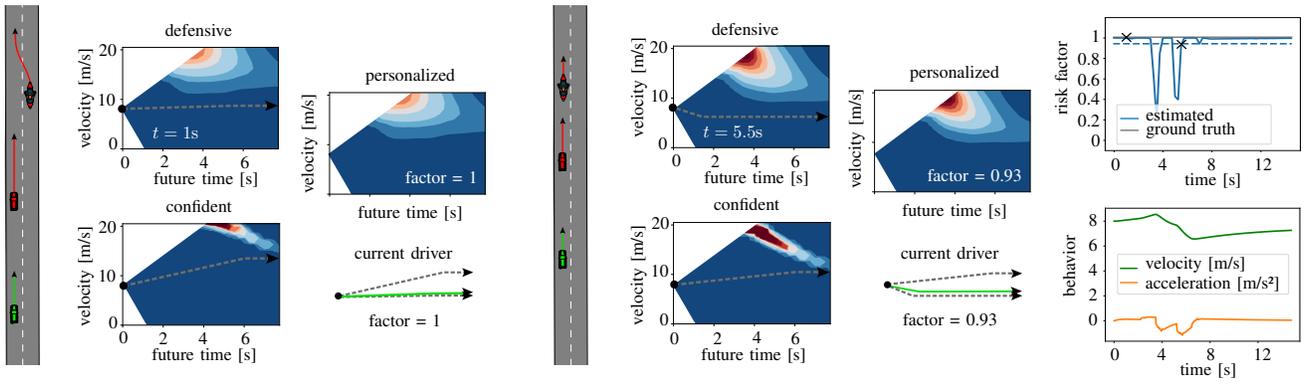}
  \vspace*{0.17cm}
  \caption[]{The image shows the risk factor estimation module applied on the car following scenario example with a defensive driver type. The driver is braking and the module thus correctly estimates the driver to be a defensive driver type. The personalized risk factor is high for the driver. The dashed line in the risk factor plot (top right) represents the average of the estimations over the whole simulation time.} 
  \label{fig:following_detailed}
\end{figure*}

\begin{figure}[t!]
  \centering
  \vspace*{0.1cm}
  \resizebox{0.9\linewidth}{!}{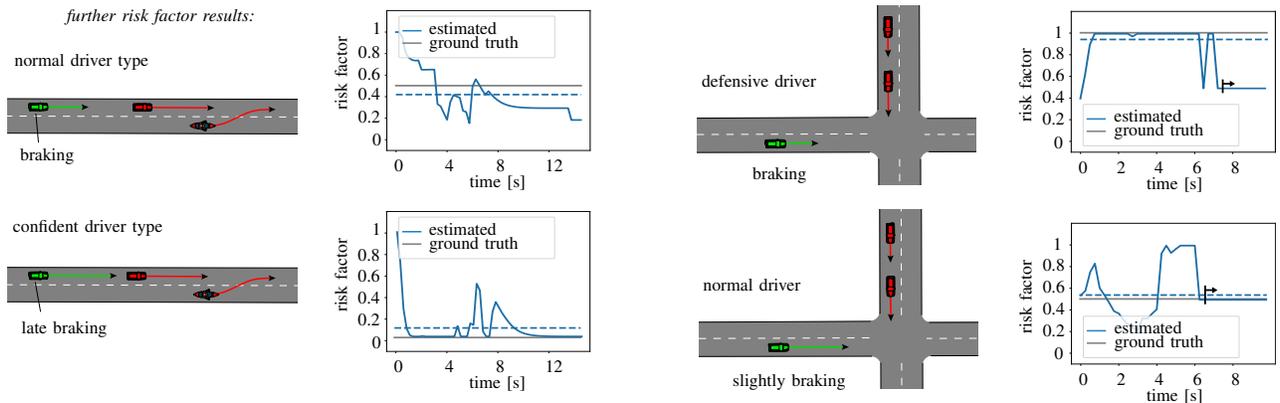}
  \caption[]{The image shows examples of the risk factor estimation results for a normal driver and a confident driver for the car-following scenario. The estimation module allows to also estimate here the risk factor.}
  \vspace{0.03cm}
  \label{fig:estimation_further_examples}
\end{figure}

\subsection{Tests for Risk Factor Estimation}
The task in the first set of tests of this paper is to correctly estimate the risk factor of the driver type in the simulation. For this purpose, the ground truth for the driver type was set in the simulation to $\alpha_{\text{def}}= 1.0$ for a defensive driver, $\alpha_{\text{norm}}=0.5$ for a normal driver and $\alpha_{\text{conf}}=0.04$ for a confident driver. The estimation module should now estimate the ground truth value for each driver.

The detailed simulation results of a defensive driver type for the following scenario example from the introduction of this paper are given in Fig. \ref{fig:following_detailed}. The defensive driver in the green car is braking because of a red car in front. The figure shows birds-eye views of the simulation and the plots of defensive and confident Risk Maps for the timesteps $t = \unit[1] {s}$ and $t = \unit[5.5] {s}$. While defensive Risk Maps plans a braking driver behavior in this scenario, confident Risk Maps plans an acceleration behavior. The interpolation between both Risk Maps shows that the risk factor is correctly estimated to be close to the high value of $\alpha_{\text{def}}=1$ in both timesteps because the actual driver is also braking. On the top right of the figure, the estimated risk factor is additionally plotted compared to the ground truth value. As can be seen, although there are jumps in the estimation, the estimated risk factor matches on average the ground truth over the whole simulation time, see dashed blue line for the average of the estimated values.

Fig. \ref{fig:estimation_further_examples} shows similar results of the estimation module for the same car-following scenario driven by a normal driver and a confident driver. In the simulation, the normal driver brakes and the confident driver accelerates for some time and brakes later. The figure shows that the risk factor estimation module can also estimate on average the risk factor for normal drivers (estimation is close to $\alpha_{\text{norm}}=0.5$) and confident drivers (estimation is close to $\alpha_{\text{conf}}=0.04$). In the estimation plots, some jumps are observable as for the confident driver simulation.

\begin{figure}[t!]
  \centering
  \vspace*{0.02cm}
  \resizebox{0.9\linewidth}{!}{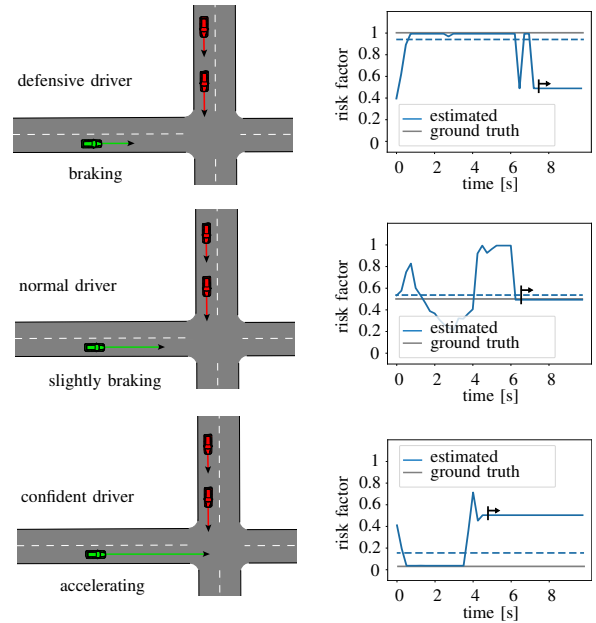}
  \vspace{0.01cm}
  \caption[]{Further examples for the risk factor estimation results. The image shows examples for a defensive driver, a normal driver and a confident driver for an intersection scenario. For all three driver types, the risk factor estimation can approximate the ground truth of the risk factor.}
  \label{fig:estimation_further_examples_intersection}
\end{figure}

The simulation results of the experiments for an intersection scenario are depicted in Fig. \ref{fig:estimation_further_examples_intersection}. The intersection scenario shows a driver in a green car that intends to cross while two red cars are approaching the intersection at the same time. Depending on the driver type, the driver can decide whether to brake and cross second or to accelerate and cross first. As depicted with the risk factor estimation plots, the estimation module also allows here to correctly approximate the ground truth of the risk factor for the defensive driver, normal driver and confident driver in intersection scenarios. Note in the figure the start of a time interval marked with a black arrow. In this interval, there is no interaction anymore between the driver to another vehicle and the risk factor becomes undetermined. The estimation module assumes in these cases a risk factor of $\alpha_{\text{norm}} = 0.5$ as for the normal driver type.  

In the estimation module tests, we averaged over all scenario tests for each driver type to retrieve the final driver risk factor estimations. These identified averaged values are used for the personalized warning system. This is similar to the practice of recording the driving of a user multiple times and then activating the personalized driver support. The averaged results are summarized in Tab. \ref{tab:averaged_values}, in which we summarize difference and standard deviation to the ground truth values. As can be seen, the maximum variation in the estimation is found for the normal driver with a standard deviation $\sigma_{\text{norm}} = 0.06$ in the risk factor.

\begin{table}[t!]
\vspace{0.15cm}
\centering
\begin{tabular}{ |p{1.2cm}|p{1.0cm}|p{1.15cm}|p{1.15cm}|p{1.15cm}|}
 \hline
 & ground truth & average & difference & standard deviation \\
 \hline
 defensive & 1.0 & 0.95 & 0.05 & 0.02 \\
 \hline
 normal & 0.5 & 0.48 & 0.02 & 0.06 \\
 \hline
 confident & 0.04 & 0.13 & 0.09 & 0.03 \\
 \hline
\end{tabular}
\caption{Results of averaging the estimated risk factor values over all driving scenario tests.}
\vspace{-0.0cm}
  \label{tab:averaged_values}
\end{table}

\subsection{Tests of Warning Errors}

In the following, we lastly present the results of using the aforementioned averaged personalized risk factor values and applying the novel warning system of personalized Risk Maps. The task in the second set of tests of this paper was to correctly warn the driver depending on the driver type.

\begin{figure*}[t!]
  \centering
  \vspace*{0.13cm}
  \resizebox{0.92\linewidth}{!}{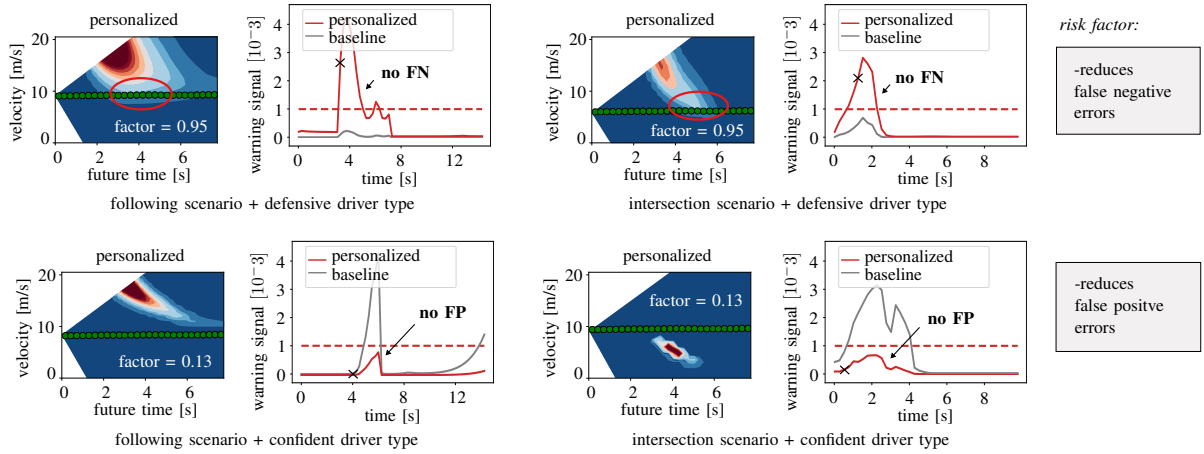}
  \vspace*{-0.1cm}
  \caption[]{Examples of improvements with personalized Risk Maps in longitudinal following and intersection scenarios. For defensive and confident driver types, warning errors can be reduced compared to the baseline approach without personalization. Defensive drivers will receive fewer false negative errors FN (see top row of image) and confident drivers will receive fewer false positive errors FP (see bottom row). In the figure, the black cross marker in the warning signal plots denotes the time for which the personalized Risk Maps plot is shown. The Risk Maps plots highlight the risk only for the possible collision with the most important other vehicle in the driving situation.}
  \label{fig:warning}
\end{figure*}

\begin{figure*}[t!]
  \centering
  \vspace*{0.07cm}
  \resizebox{0.92\linewidth}{!}{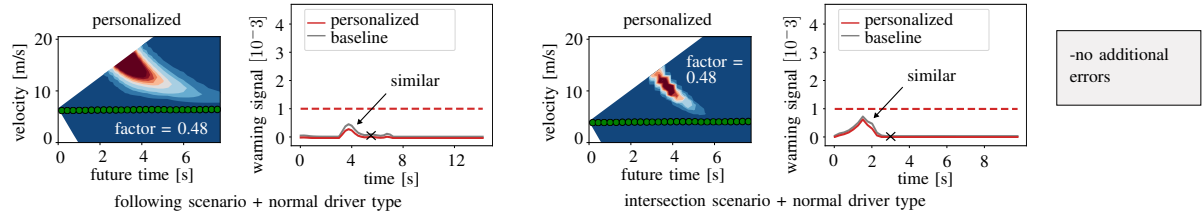}
  \vspace*{-0.15cm}
  \caption[]{Examples of no improvements with personalized Risk Maps. For normal drivers, personalization does not reduce warning errors because the parametrization is similar to the baseline approach without personalization.}
  \vspace*{-0.09cm}
  \label{fig:warning2}
\end{figure*}

For the risk warning system, we first set the risk weights to $w_{\text{def}} = 4.0$ for the defensive driver, $w_{\text{norm}} = 1.0$ for the normal driver and $w_{\text{conf}} = 0.01$ for the confident driver. In order to show improvements, we then compared the novel warning system with the baseline approach \cite{puphal2022} that does not use a personalized model. The baseline approach always evaluates here the risk with a normal driver parametrization and uses this as a warning signal. When the risk now exceeds the risk threshold $R_{\text{thr}} = 10^{-3}$ for one of the approaches in the simulation, a warning for the respective approach was recorded. The warnings of the personalized system could thus be compared with the baseline approach.

The improvements of personalized Risk Maps in the tests of warning errors are shown in 
Fig. \ref{fig:warning}. As is evident, for defensive and confident driver types, warning errors can be reduced compared to the baseline approach. For a defensive driver, false negative warning errors FN are avoided both in following and intersection scenarios (compare top row of figure). 
As an example, the proposed system warns the defensive driver at around $t=\unit[3] {s}$ in the following scenario. The constant driver behavior prediction goes over the large risk spot of personalized Risk Maps with the used estimated risk factor $\tilde{\alpha}_{\text{def}} = 0.95$. In comparison, the warning signal of the baseline approach does not exceed the warning threshold $R_{\text{thr}}=10^{-3}$ in the simulation run. The same improvement can be achieved also for the intersection case. The false negative error FN is reduced by using the proposed personalized system. 

For a confident driver, false positive warning errors FP can additionally be reduced in the simulated scenarios (compare bottom row of the figure). A confident driver does not like to get warned for the given medium-risk driving situations. Personalized Risk Maps correctly, in this case, does not warn the driver and avoids a false positive error FP in the examples of car-following and intersection driving. A reduced risk value for different driver behaviors can be seen by the small risk spots in the personalized Risk Maps plot. In comparison, the baseline approach exceeds the risk threshold and warns the driver in these medium-risk situations.

Finally, Fig. \ref{fig:warning2} shows examples of no improvements with the novel warning system of personalized Risk Maps. For normal drivers, the personalization of the warning system matches the general normal driver parametrization of the baseline approach. Therefore, personalized Risk Maps has the same behavior as the baseline approach. The Risk Maps plot is similar to the baseline and thus the warning signal is also similar. As depicted in the figure, both systems do not warn in the longitudinal following and intersection scenario of the example simulations. In the case of a normal driver, personalization therefore cannot reduce warning errors.

\subsection{Discussion}
Overall, the results showed that personalized Risk Maps allows to effectively reduce warning errors in some cases for driver support compared to the baseline approach not using personalized models. Improvements were presented for examples in both car-following and intersections scenarios. This highlights the generalizability of the personalized risk warning system in comparison to previous works in literature. However, there are some limitations of the experiment results. In this last section, we will shortly discuss these limitations and explain possible improvements for future work.

As could be seen in the risk factor estimation tests, the risk factor estimation contained jumps. Since a simple interpolation is used for the estimation module, the module can only classify the three discrete driver types of defensive, normal and confident drivers. Here, the normal driver type was the most difficult class to estimate because the interpolation does not use a third reference point for the risk factor located at the normal driver parametrization. More complex estimation methods than interpolation could improve here the risk factor estimation capabilities.
 
Another limitation in the experiments was that the same behavior planner Risk Maps was used in the simulation as for the risk estimation and warning system. The tests were not done with real driver parameters and a real driver could exhibit a more complex behavior than defensive, normal and confident driver behaviors. In order to validate the results of this paper, a user study in a real vehicle is needed that investigates the possible user experience improvement with personalized Risk Maps. With these limitations in mind, the results showcase promising improvements with personalization for general risk warning.

\vspace{0.05cm}\section{Conclusion and Outlook}
\label{sec:conclusion}

In summary, in this paper, we proposed a novel driver support in the form of general personalized risk warning. The system estimates a personalized risk factor of the driver based on the sensed driver behavior and uses the factor in personalized Risk Maps. Consequently, the system could distinguish and leverage defensive, normal and confident driver types for improving driver risk warning. In comparison, previous work focused more on vehicle control systems and driver models and only few approaches were proposed that use personalized models for warning. 

In experiments of variations for longitudinal following and intersection scenarios, we finally showed that the estimation module can estimate personalized risk factors in simulation. In effect, personalized Risk Maps was able to reduce false negative warning errors and false positive warning errors in some cases of the scenario simulations. The results highlight the potential of using personalization for reducing warning errors in general risk warning. Previous warning systems were limited in the applications they handled (i.e., lane departure warning or frontal collision warning). We hope that this paper therefore motivates further research in improved general driver risk warning.

For future work, we are planning to improve the risk \makebox[\linewidth][s]{warning system and apply the system on a prototype vehicle}

\newpage

\noindent in order to validate the results of reducing warning errors. The chosen Human-Machine Interface (HMI) for the risk warning can affect here the outcome of a possible user study and should be considered. Different modalities, such as audio or haptic feedback, may be used depending on the driver type. As an example, the work of \cite{krueger2023} investigated various modalities for warning of frontal risks and could be combined with the personalized models of this paper.

Moreover, driver-machine interaction methods that use the personalized Risk Maps are promising for further improvements. The risk factor is in the current proposed system assumed to be constant for each driver. We target to improve the system so that the system can slowly adapt the risk factor online during many drives in order to educate the driver to drive with a target driving type. In this way, a confident driver can be motivated, for example, to drive similarly to a normal driver type. The general driver safety may thus be enhanced with such an interactive driver warning system.

\addtolength{\textheight}{-12cm}   
              
\bibliographystyle{IEEEtran}
\bibliography{bib}

\end{document}